\documentclass{article}

\usepackage[final]{tackling_climate_workshop_style}

\usepackage[utf8]{inputenc} 
\usepackage[T1]{fontenc}    
\usepackage{hyperref}       
\usepackage{url}            
\usepackage{booktabs}       
\usepackage{amsfonts}       
\usepackage{nicefrac}       
\usepackage{microtype}      

\usepackage{float}
\usepackage{multirow}
\usepackage{threeparttable}
\usepackage{graphicx}

\title{Soil Organic Carbon Estimation from Climate-related Features with Graph Neural Network}

\author{%
  Weiying Zhao \\
  Deep Planet \\
  London, UK \\
  \texttt{weiying@deepplanet.ai} \\
  \And
  Natalia Efremova\\
  Queen Mary University and the Alan Turing Institute \\
  London, UK \\
  \texttt{natalia.efremova@gmail.com} \\
}

\begin{document}

\maketitle

\begin{abstract}
Soil organic carbon (SOC) plays a pivotal role in the global carbon cycle, impacting climate dynamics and necessitating accurate estimation for sustainable land and agricultural management. While traditional methods of SOC estimation face resolution and accuracy challenges, recent  technological solutions harness remote sensing, machine learning, and high-resolution satellite mapping. Graph Neural Networks (GNNs), especially when integrated with positional encoders, can capture complex relationships between soil and climate. Using the LUCAS database, this study compared four GNN operators in the positional encoder framework. Results revealed that the PESAGE and PETransformer models outperformed others in SOC estimation, indicating their potential in capturing the complex relationship between SOC and climate features. Our findings confirm the feasibility of applications of  GNN architectures in SOC prediction, establishing a framework for future explorations of this topic with more advanced GNN models.
\end{abstract}

\section{Introduction}

Soil organic carbon plays major role in the global carbon cycle, acting as both a source and sink of carbon and profoundly influencing soil health, fertility, and overall ecosystem functionality. Accurate estimation and monitoring of SOC is crucial for understanding climate dynamics and driving sustainable land management and agricultural practices. Climate change profoundly impacts SOC dynamics by influencing various processes related to plant growth, microbial activity, and organic matter decomposition. Traditional methods of SOC estimation often face challenges in spatial resolution, coverage, and accuracy, particularly when applied at larger scales or diverse landscapes [1].
Recent technological advancements brought SOC monitoring to a new level, employing methods like remote sensing, machine learning, and satellite-driven high-resolution mapping [2,3].  Satellite imagery offers a scalable and cost-effective solution, capturing spatial heterogeneity, temporal dynamics of SOC, and variability of climate features across regions, from South Africa's diverse landscapes [4] to Bavaria's agriculturally intensive zones [5].

Graph Neural Networks have the ability to model complex interdependencies between SOC and multifaceted climate features. By design they excel at capturing relational information in data [6], which makes them indespansable for modelling the relationship between soil and climate. For instance, integrating positional encoders [7] in GNNs allows them to capture the spatial dependencies crucial for geographic data. Furthermore, GNNs that learn both structural and positional representations provide a comprehensive understanding, ensuring that elements' composition and spatial distribution are considered. In this paper, we propose applying advanced GNN operators [8-11] to the positional encoder framework [7] for SOC estimation. We aim to use their relational modelling capabilities and computational optimizations to deliver accurate, detailed, and scalable SOC predictions across diverse climate-related features.

\section{Methodology}

\begin{figure}[H]
  \centering
  \includegraphics[width=0.95\textwidth]{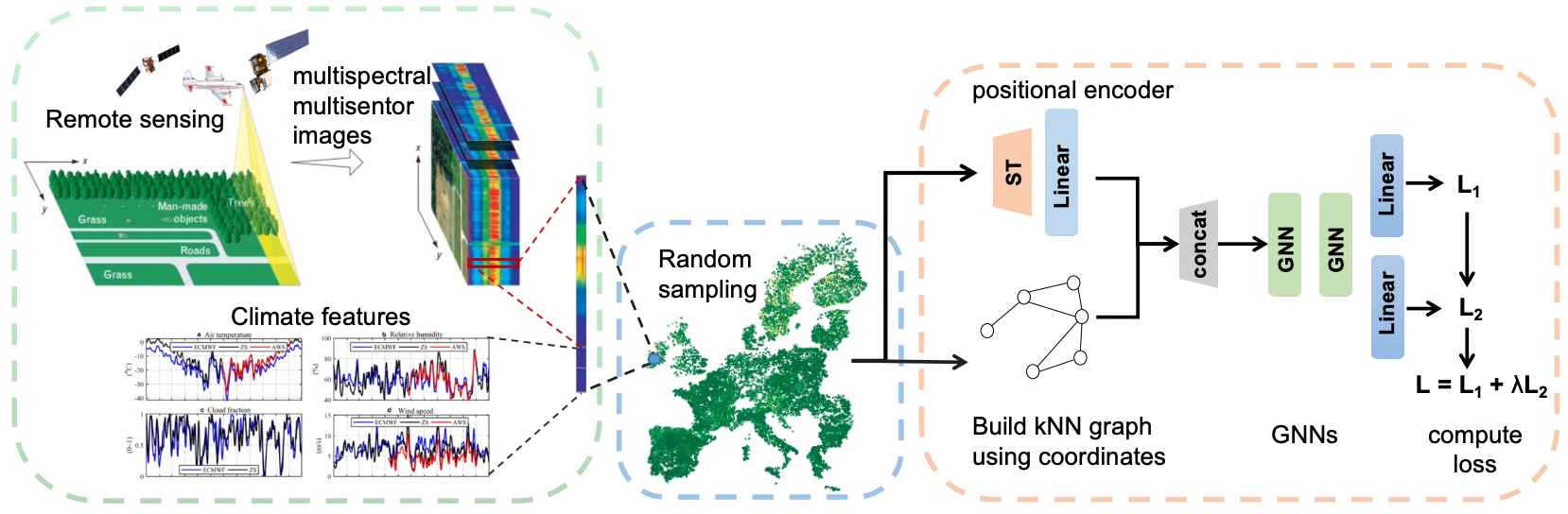}
  \caption{The entire process of SOC estimation with PE-GNN. PE-GNN contains a (1) positional encoder network, learning a
spatial context embedding and (2) an auxiliary learner, predicting the spatial autocorrelation of the outcome variable simultaneously to the main regression task [7].}
  \label{fig:pegcn_flow}
\end{figure}

\subsection{Message passing networks}

In graph-based learning, the challenge lies in adapting traditional convolution operators, which thrive on regular grid structures, to work effectively on irregular graph domains. One widely adopted approach to address this challenge is through the notion of message passing or neighbourhood aggregation. In this context, let's define \( x^{(k)}_i \in \mathbb{R}^F \) as the features associated with node \( i \) at the $kth$ layer, \( e_{j,i} \in \mathbb{R}^D \) represents the edge features from node \( j \) to node \( i \). Given this, the operation of message-passing  can be articulated as:

\quad
\begin{equation}
    \mathbf{x}^{(k)}_i = \gamma^{(k)} \left( \mathbf{x}^{(k-1)}_i,
        \bigoplus_{j \in \mathcal{N}(i)} \, \phi^{(k)}
        \left(\mathbf{x}^{(k)}_i, \mathbf{x}^{(k-1)}_j,\mathbf{e}_{j,i}\right) \right)
\end{equation}

where \( \bigoplus \) is a differentiable function that remains invariant to permutations. Common choices for this function include operations like sum, mean, or max. The functions \( \gamma \) and \( \phi \) are differentiable mappings, often realized using structures like Multi-Layer Perceptrons (MLPs).

With this scheme, we proceed to evaluate four prominent operators in the positional encoder framework: GCN, SAGE, Transformer [11], and GAT (Tab. \ref{sample-table}).

\begin{table}[H]
  \caption{Message passing operators}
  \label{sample-table}
  \centering
  \begin{tabular}{ll}
    \toprule
    operator&	equation\\
    \midrule
    \multirow{2}{*}{GCNConv} & $\mathbf{x}^{\prime}_i = \mathbf{\Theta}^{\top} \sum_{j \in
        \mathcal{N}(i) \cup \{ i \}} \frac{e_{j,i}}{\sqrt{\hat{d}_j
        \hat{d}_i}} \mathbf{x}_j$ \\
     & $\hat{d}_i = 1 + \sum_{j \in \mathcal{N}(i)} e_{j,i}$ \\
    \midrule
    SAGEConv&	$\mathbf{x}^{\prime}_i = \mathbf{W}_1 \mathbf{x}_i + \mathbf{W}_2 \cdot
        \mathrm{mean}_{j \in \mathcal{N}(i)} \mathbf{x}_j$\\
    \midrule
    \multirow{2}{*}{TransformerConv} & $\mathbf{x}^{\prime}_i = \mathbf{W}_1 \mathbf{x}_i +
        \sum_{j \in \mathcal{N}(i)} \alpha_{i,j} \mathbf{W}_2 \mathbf{x}_{j}$ \\
     & $\alpha_{i,j} = \textrm{softmax} \left(
        \frac{(\mathbf{W}_3\mathbf{x}_i)^{\top} (\mathbf{W}_4\mathbf{x}_j)}
        {\sqrt{d}} \right)$ \\
    \midrule

    \multirow{2}{*}{GATConv} & $\mathbf{x}^{\prime}_i = \alpha_{i,i}\mathbf{\Theta}\mathbf{x}_{i} +
        \sum_{j \in \mathcal{N}(i)} \alpha_{i,j}\mathbf{\Theta}\mathbf{x}_{j}$ \\
     & $\alpha_{i,j} =
        \frac{
        \exp\left(\mathrm{LeakyReLU}\left(\mathbf{a}^{\top}
        [\mathbf{\Theta}\mathbf{x}_i \, \Vert \, \mathbf{\Theta}\mathbf{x}_j]
        \right)\right)}
        {\sum_{k \in \mathcal{N}(i) \cup \{ i \}}
        \exp\left(\mathrm{LeakyReLU}\left(\mathbf{a}^{\top}
        [\mathbf{\Theta}\mathbf{x}_i \, \Vert \, \mathbf{\Theta}\mathbf{x}_k]
        \right)\right)}$ \\
    \bottomrule
    \multicolumn{2}{l}{Note: $\alpha_{i,j}$ is the attention coefficients, $\mathbf{W}$ represents the weight.}\\
  \end{tabular}
\end{table}

\subsection{Positional encoder graph neural network}

Unlike the standard GNN method, PE-GNN (Fig.\ref{fig:pegcn_flow}) integrates a positional encoder that transforms 2D geographic coordinates into context-aware vector embeddings. This allows for a flexible representation of spatial context and relationships. A spatial graph is constructed for each batch using the k-nearest neighbours method in the training process. The outcome variable's local Moran’s I values, an autocorrelation metric, are computed, generating a "shuffled" version of the metric due to randomized minibatching. The PE-GNN model uses two prediction heads with shared graph operation layers, and its loss calculation incorporates both the main task and the auxiliary Moran’s I task, weighted by a parameter $\lambda$. This unique design enables PE-GNN to learn spatial complexities in a more adaptable manner, considering relationships between varying clusters of points across iterations. Consequently, this helps the model generalize better and not rely on memorized neighbourhood structures. Combining the positional encoder framework with the previous four operators, we got PEGCN, PESAGE, PETransformer and PEGAT.

\section{Results and Discussion}

We compared the four operators using the Land Use/Land Cover Area Frame Survey (LUCAS) database [12], a harmonised in situ land cover and land use data collection  over the whole of the EU’s territory. The terrain attributes are derived from COPERNICUS and USGS DEMs. Macroclimate features come from ERA5-Land Daily Aggregated data (ECMWF). Landsat-8 offers high-resolution landcover images, MODIS supplies medium-resolution surroundings, and OpenLandMap contributes some soil attributes. All the data are prepared with the help of Google Earth Engine. 
The prepared dataset, comprising 21,245 samples with 42 features each, was collected from  cropland and grassland in 2015 and 2018. All methods shared a comparison framework. Given the target value's heavy-tailed distribution, we log-transform it to mitigate outlier effects and expedite training.
The data split was 70\% training, 15\% testing, and 15\% evaluation. 

\begin{figure}[H]
\graphicspath{{results/}}
   \centering
\begin{tabular}{cccc}
\multicolumn{2}{c}{\includegraphics[width=5.2cm]{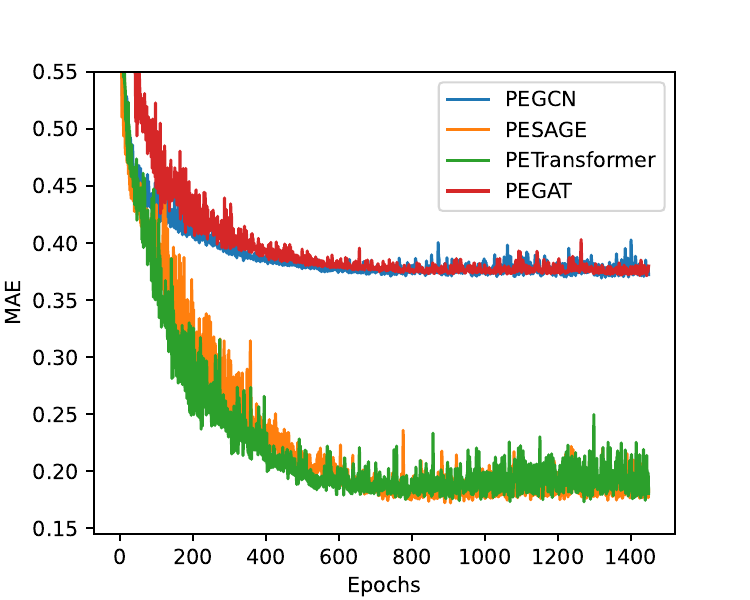}}&
\multicolumn{2}{c}{\includegraphics[width=5.2cm]{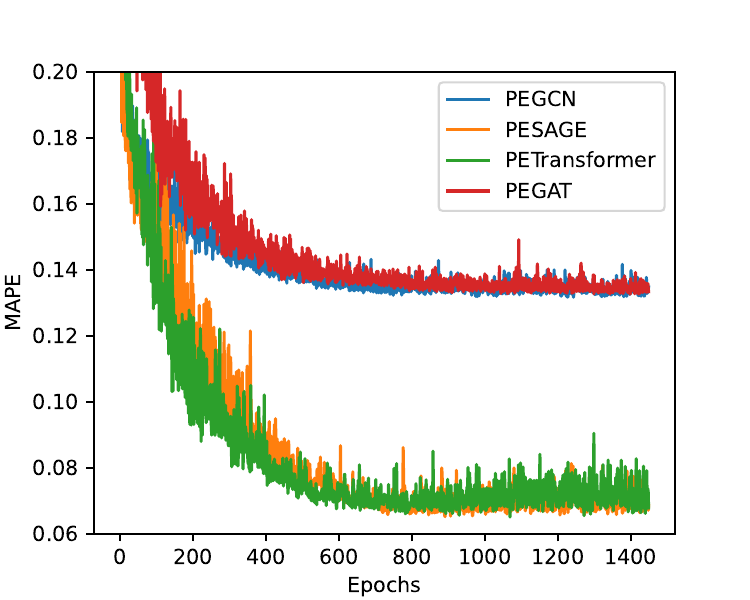}}\\
 \multicolumn{2}{c}{(a) MAE}& \multicolumn{2}{c}{(b) MAPE}\\
\end{tabular}
  \caption{Test error curves of PEGCN, PESAGE, PETransformer and PEGAT models, measured by the MAE and MAPE metrics.}
  \label{fig:training_performance} 
\end{figure}

According to Fig.\ref{fig:training_performance}, PESAGE and PETransformer methods provide better testing performance than others during the training. 
Since climate features have different dimensions and scales, for each point,  we convert all of them to a single vector. It'll make the features have complex distributions, which is not good for GCNConv. 

GATConv incorporates attention mechanisms to weigh neighbour contributions. In testing with head=1, the single attention mechanism may limit concurrent focus on diverse graph regions. SAGEConv and TransformerConv might be capturing more complex patterns in the data due to their aggregation mechanisms and self-attention capabilities, respectively. The training process, including learning rates, regularization, and other hyperparameters, can also impact the performance of these layers. It's possible that the training setup was more favorable for SAGEConv and TransformerConv. The evaluation results shown in Tab.\ref{spatial_interpolation} also proved this. PESAGE with $\lambda$ equals 0.5 provides the best testing results.

\begin{figure}[H]
  \centering
  \includegraphics[width=1.\textwidth]{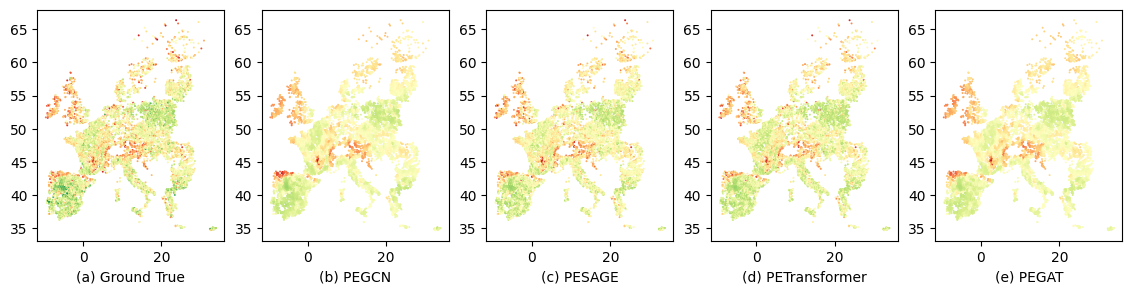}
  \caption{Spatial variance of ground truth data measured in 2018 and predicted SOC provided by different methods. Green represents lower SOC values, while red represents higher values.}
  \label{fig:evaluation_spatial_variance_18}
\end{figure}

\begin{figure}[H]
  \centering
  \includegraphics[width=0.95\textwidth]{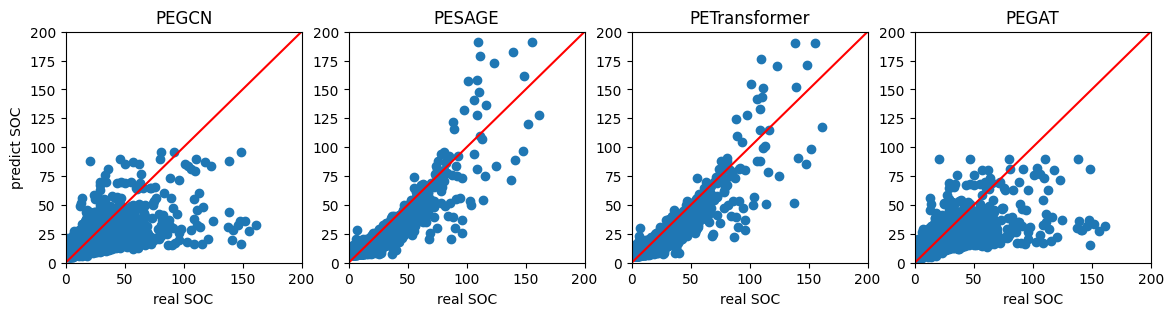}
  \caption{Scatter plot of real SOC and predicted SOC provided by different methods on the validation database.}
  \label{fig:scatter_plot_eval_part}
\end{figure}

Compared with the ground truth data, the spatial variance of the predicted values shown in Fig.\ref{fig:evaluation_spatial_variance_18} provided by PEGCN and PEGAT have been smoothed, while the other methods can provide fine features. The fundamental operation of graph convolutional networks  is to aggregate information from neighbouring nodes. In GCNConv, node features are aggregated using a simple weighted average of their neighbours. This kind of aggregation tends to produce a smoothing effect over the graph.
While GATConv introduces an attention mechanism that weighs the contributions of neighbouring nodes, the attention weights can sometimes lead to a kind of averaging, especially if the attention scores do not vary significantly among the neighbours.  One of the methodologies in SAGEConv is to sample a fixed-size set of neighbors at each layer. This sampling can prevent the rapid expansion of receptive fields, thereby reducing the over-smoothing effect seen in traditional GCNs. SAGEConv often concatenates the current node's features with aggregated neighbor features, helping to preserve the node's original information.  TransformerConv use positional encodings, which could add more distinctiveness to node embeddings, reducing the chances of over-smoothing. The transformer has residual connections, which can help retain original information and prevent over-smoothing by allowing gradients to flow directly through layers. PESAGE and PETransformer provide better evaluation results as shown in Fig.\ref{fig:scatter_plot_eval_part}.

\begin{table}[H]
  \caption{Spatial interpolation performance based on log-transformed target values: Test MSE, MAE and MAPE scores using four GNN backbones with position encoder architecture. Data split 70/15/15.}
  \label{spatial_interpolation}
  \centering
  \small
  \begin{tabular}{lllll}
    \toprule
    Method&	$\lambda$&	MSE&	MAE& MAPE\\
    \midrule
PEGCN&	0.25&	0.2464±0.0049&	0.3786±0.0049&	0.1356±0.0016\\
PEGCN&	0.5&	0.2449±0.0059&	0.3775±0.0055&	0.1346±0.0018\\
PEGCN&	0.75&	0.2457±0.0045&	0.3783±0.0049&	0.1346±0.0009\\
\midrule
PEGraphSAGE&	0.25&	0.0750±0.0101&	0.1968±0.0164&	0.0734±0.0054\\
PEGraphSAGE&	0.5&	\textbf{0.0738±0.0066}&	\textbf{0.1947±0.0113}&	\textbf{0.0727±0.0037}\\
PEGraphSAGE&	0.75&	0.0800±0.0106&	0.2048±0.0159&	0.0760±0.0052\\
\midrule
PETransformer&	0.25&	0.0814±0.0118&	0.2066±0.0171&	0.0770±0.0058\\
PETransformer&	0.5&	0.0763±0.0116&	0.1988±0.0189&	0.0743±0.0062\\
PETransformer&	0.75&	0.0782±0.0087&	0.2022±0.0138&	0.0754±0.0049\\
\midrule
PEGAT&	0.25&	0.2492±0.0079&	0.3807±0.0074&	0.1352±0.0018\\
PEGAT&	0.5&	0.2499±0.0067&	0.3817±0.0065&	0.1352±0.0018\\
PEGAT&	0.75&	0.2497±0.006&	0.3810±0.0059&	0.1355±0.0019\\
    \bottomrule
  \end{tabular}
\end{table}

\section{Conclusion}

This study emphasized the significance of SOC estimation within the global carbon cycle and its complex relationship with climate dynamics. By leveraging the LUCAS database, we explored the potential of GNNs, particularly the PESAGE and PETransformer models, in addressing the challenges faced by traditional SOC estimation methods. Our findings showed that GNN architectures can capture the complex interdependencies between SOC and climate-related features, setting a new benchmark in SOC prediction. Based on these insights, in future research we will explore the GPS graph transformer [13] to enhance SOC prediction methods.

\section*{References}
\medskip
\small

[1] Anthony D Campbell, et al.(2022) { \it A review of carbon monitoring in wet carbon systems using remote sensing}. Environmental Research Letters, 17(2):025009, 2022.

[2] Camile Sothe et al. (2022) { \it  Large scale mapping of soil organic carbon concentration with 3d machine learning and satellite observations}. Geoderma, 405:115402.

[3] Ken CLWong, et al. (2022) { \it Image-based soil organic carbon estimation from multispectral satellite images with fourier neural operator
and structural similarity}. In NeurIPS 2022 Workshop on Tackling Climate Change with Machine Learning.

[4] Zander S Venter, et al. (2021) { \it Mapping soil organic carbon stocks and trends with satellite-driven high resolution maps over South Africa}. Science of the Total Environment, 771:145384.

[5] Simone Zepp,et al. (2023) {\it Optimized bare soil compositing for soil organic carbon prediction of topsoil croplands in bavaria using
landsat}. ISPRS Journal of Photogrammetry and Remote Sensing, 202:287–302, 2023.

[6] Sergi Abadal et al. (2021) {\it Computing graph neural networks: A survey from algorithms to accelerators}. ACM Computing Surveys (CSUR), 54(9):1–38.

[7] Konstantin Klemmer, et al. (2023) { \it Positional encoder graph neural networks for geographic data}. In International Conference on Artificial Intelligence and Statistics, pages 1379–1389.

[8] Thomas N Kipf and Max Welling. {\it Semi-supervised classification with graph convolutional networks}.
arXiv preprint arXiv:1609.02907, 2016.

[9] Will Hamilton, Zhitao Ying, and Jure Leskovec. {\it Inductive representation learning on large graphs.} Advances in neural information processing systems, 30.

[10] Petar Veličković, et al. (2018) { \it Graph attention networks}. ICLR 2018.

[11] Yunsheng Shi, et al. (2020) { \it Masked label prediction: Unified message passing model for semi-supervised classification}. 30th International Joint Conference on Artificial Intelligence (IJCAI-21).

[12] Raphaël d’Andrimont, et al. (2020) { \it Harmonised LUSAS in-situ land cover and use database for field surveys from 2006 to 2018 in the european union}. Scientific data, 7(1):352.

[13] Ladislav Rampášek et al. (2022) { \it Recipe for a general, powerful, scalable graph transformer}. Advances in Neural Information
Processing Systems, 35:14501–14515.
\end{document}